\documentclass[a4paper,review]{cas-dc}

\usepackage[authoryear]{natbib}
\usepackage{tabularx}
\usepackage{caption}
\usepackage{amsfonts,amssymb,amsmath,amsgen,amsopn,amsbsy,theorem,graphicx,epsfig}

\def\tsc#1{\csdef{#1}{\textsc{\lowercase{#1}}\xspace}}
\tsc{WGM}
\tsc{QE}
\tsc{EP}
\tsc{PMS}
\tsc{BEC}
\tsc{DE}

\begin{document}\sloppy
\let\WriteBookmarks\relax
\def\floatpagepagefraction{1}
\def\textpagefraction{.001}
\shorttitle{Evaluation of Neural NER Models on Turkish }
\shortauthors{Aras et~al.}

\title [mode = title]{An Evaluation of Recent Neural Sequence Tagging Models in Turkish Named Entity Recognition}                      


\author[1]{Gizem Aras}
\cormark[1]
\ead{gizem.aras@demirorenteknoloji.com}

\address[1]{Demiroren Teknoloji A.S., Istanbul, Turkey}

\author[1,2]{Didem Makaroglu}
\ead{makaroglu17@itu.edu.tr}

\author[3]{Seniz Demir}[orcid=0000-0003-4897-4616]
\ead{demirse@mef.edu.tr}

\address[2]{Department of Physics Engineering, Istanbul Technical University, Istanbul, Turkey}

\author[2]{Altan Cakir}[orcid=0000-0002-8627-7689]
\ead{altan.cakir@itu.edu.tr}

\address[3]{Department of Computer Engineering, MEF University, Istanbul, Turkey}

\cortext[cor1]{Corresponding author}

\begin{abstract}
Named entity recognition (NER) is an extensively studied task that extracts and classifies named entities in a text. NER is crucial not only in downstream language processing applications such as relation extraction and question answering but also in large scale big data operations such as real-time analysis of online digital media content. Recent research efforts on Turkish, a less studied language with morphologically rich nature, have demonstrated the effectiveness of neural architectures on well-formed texts and yielded state-of-the art results by formulating the task as a sequence tagging problem. In this work, we empirically investigate the use of recent neural architectures (Bidirectional long short-term memory and Transformer-based networks) proposed for Turkish NER tagging in the same setting. Our results demonstrate that transformer-based networks which can model long-range context overcome the limitations of BiLSTM networks where different input features at the character, subword, and word levels are utilized. We also propose a transformer-based network with a conditional random field (CRF) layer that leads to the state-of-the-art result (95.95\% f-measure) on a common dataset. Our study contributes to the literature that quantifies the impact of transfer learning on processing morphologically rich languages.

\end{abstract}

\maketitle

\section{Introduction}
Named entity recognition (NER) aims to recognize named entities in a given text by determining their boundaries and classifying them into predefined categories (e.g., person, location, and temporal expression). NER is a crucial step in various natural language processing applications such as event extraction~\citep{Chen:15} and question answering~\citep{Molla:06} as well as in big data analytics~\citep{Saju:17}. Early studies have addressed the recognition of named entities as a sequence labeling problem and extensive research efforts have been devoted to developing solutions using machine learning techniques~\citep{Lin:06, Ekbal:08}, hidden markov models~\citep{Zhou:02}, and conditional random fields~\citep{Yao:09, Zirikly:15}. Recently, neural models have been introduced to named entity task in well-formed and noisy texts~\citep{Alnabki:20}. In spite of recent advances, NER remains to be a challenging problem due to several reasons such as the recognition of overlapping or nested entities, infrequent entities in user generated noisy texts, and semantically ambiguous entities in different contexts.

In the current era, the amount of online content has exploded which makes it exhaustive to search from a vast distributed source of information. Search tools or expert systems might effectively alleviate the problem of accessing available content on the web. However, continuous alteration of natural languages due to heavy social media usage, social-cultural factors in society, daily events (e.g., political changes and major sport events) has reflections in written texts and leads to constant evolution of words, expressions and importantly named entities. Correctly identified named entities from unstructured or semi-structured content form a basis for the development of more effective and intelligent information management, text mining, and relation extraction systems~\citep{Marrero:13}. For instance, mining daily news content by digital media applications for extracting information about a person or a location  necessitates querying an astonishing amount of news articles which can be facilitated by automatic detection of named entities in written texts. Paving the road for interpretable and reusable information through semantically annotated online content can also be listed as a particular benefit of extracting named entities and their relations from raw texts.

NER is a well-studied task for several languages including Turkish and recent successes in neural architectures have greatly advanced achieved performances on recognizing Turkish named entities~\citep{Gunes:18, Gungor:19}. In these studies, Bidirectional Long Short Term Memory networks with different word representations were widely used and evaluated on a common dataset consisting of person, location, and organization names~\citep{Tur:03}. A conditional random field (CRF) was shown to positively contribute to these networks that minimize the need for feature engineering. There is a recent interest in applying deep bidirectional transformers~\citep{Stefan:20} and transfer learning~\citep{Akkaya:20} to Turkish entity tagging. In this work, we present a comprehensive evaluation of two notable neural architectures, namely BiLSTM networks and Transformer-based networks and compare their performances in the same experimental setting. In BiLSTM models, we explore different combinations of four kinds of embeddings as input (i.e., character, morphological, subword, and word embeddings) and experiment with different pretrained embeddings as initializations of word embeddings. In transformer-based models, we benefit from three different transformer based language models, namely multilingual cased BERT (mBERT), Turkish BERT (BERTurk), and XLM-RoBERTa (XLMR), and study the effectiveness of both linear and CRF layers at the top of the network. As our second contribution, we propose a transformer-based neural architecture accompanied with a CRF as the top layer (an extension of the BERTurk model) which sets the new state-of-the-art f-measure of 95.95\%. Our study not only extends the current Turkish NER literature but also validates the usability of transfer learning on processing a morphologically rich language. 

The rest of this article is organized as follows. Section~\ref{sec:Literature} discusses related research on named entity recognition with a particular focus on Turkish NER studies. Section~\ref{sec:Architecture} describes neural architectures utilized in this work. Section~\ref{sec:Experiments} presents our dataset and parameter initializations used for building neural architectures. Section~\ref{sec:Results} discusses conducted experiments and the results that we obtained. Finally, Section~\ref{sec:Conclusion} concludes the article and presents our future work.

\section{Literature Review} \label{sec:Literature}

\begin{table*}[pos=h]
\caption{English NER Studies}
\begin{tabular}{llll}
\toprule
Study      & Approach   & Embedding  & F1 Score \\
\midrule
\citet{Ma:16}  & CNNChar-BiLSTM-CRF     & -                                                                                        & 80.76      \\
\citet{Collobert:11}  & Tanh-CRF       & -                                                                                 & 81.47      \\
\citet{Huang:15}    &BiLSTM-CRF       & -                                                                                         & 84.26      \\
 \citet{Huang:15}  & BiLSTM-CRF          & Senna                                                                                       & 84.74      \\
 \citet{Ma:16}  & CNNChar-BiLSTM-CRF    & Skip-Gram                                                                            & 84.91      \\
\citet{Collobert:11}  & Tanh-CRF      & Senna                                                                                             & 88.67      \\
 \citet{Huang:15}   & BiLSTM-CRF         & Senna                                                                                         & 88.83      \\
\citet{Ma:16}  & CNNChar-BiLSTM-CRF    & Senna                                                                                        & 90.28      \\
\citet{Lample:16} & LSTMChar-BiLSTM-CRF  & Skip-Gram                                                                   & 90.96      \\
\citet{Ma:16}  & CNNChar-BiLSTM-CRF     & Glove                                                                                              & 92.21      \\
\citet{Akbik:18}&Flair-Char-BiLSTM-CRF      & Glove                                                                                  & 93.09     \\ \bottomrule
\end{tabular}
\label{tabl:en_lit_rev}
\end{table*}

\subsection{Neural Models for Named-Entity Recognition} \label{subsec:Neural}
Earlier traditional named entity recognition systems have relied heavily on feature engineering and employed hand-crafted language dependent features, large gazetteers, and tagged datasets \citep{Collobert:11}. A significant branch of research has utilized a range of statistical approaches to address the problem such as maximum entropy classifiers~\citep{Chieu:03}, decision trees~\citep{Paliouras:00}, and conditional random fields~\citep{Finkel:09}. However, in recent years, the focus of NER research has shifted to neural models in parallel with observed improvements on multiple language processing benchmarks such as question-answering and language generation. Neural NER models have been guided by distributional approaches where the meaning of a word is carried in its surroundings via vector representations~\citep{Harris:54, Firth:57, Mikolov:13-2}. Initial attempts considered words as separate tokens and represented each token using a fixed-length vector~\citep{Mikolov:13, Pennington:14}. Some other studies explored different ways of representing words such as concatenating embeddings of characters~\citep{Santos:15}, morphemes~\citep{Luong:13}, or other word subparts to fixed-length word embeddings. In recent NER studies, the problem was formulated as a sequence labeling task and different Seq2Seq models were shown to achieve state-of-the art results where final embeddings of words are encoded by gated recurrent units (GRUs) or long short term memory units (LSTMs)~\citep{Lample:16, Ma:16, Chen:18}. Using conditional random fields on top of neural networks were proved to work equally well~\citep{Collobert:11, Huang:15, Chiu:16} or better than previous methods. Moreover, BiLSTM-CRF models with character and word embeddings were shown to be effective for multiple languages including Chinese~\citep{Zhang:19} and arguably considered as a base model for tagging~\citep{Jurafsky:08}. Unfortunately, a word, no matter in which context it appears, is represented with the same final embedding in these models. A recent study has utilized contextual string embeddings~\citep{Akbik:18}, where final word embeddings are contextualized according to the entire sentence. In that study, all characters in the sentence up to the last character of a word were processed via a forward LSTM and all characters from the end to the beginning of the sentence were processed via a backward LSTM. The obtained hidden states were then concatenated to produce final embedding of the word in focus. This kind of word embeddings was proved to improve not only NER tagging but also other sequence labeling tasks such as part-of-speech labeling and phrase chunking. Achieved performance scores (f1 scores) of some these English NER studies are given in Table~\ref{tabl:en_lit_rev}.

Transformer-based approaches outperformed the state-of-the-art on several NLP tasks and achieved performance improvements that might be attributed to the use of attention-mechanism~\citep{Vaswani:17}. Bidirectional Encoder Representations from Transformers (BERT)~\citep{Devlin:18} is a bi-directional transformer that learns contextualized input representations. BERT is different from earlier work in four aspects. First, it uses transformers \citep{Vaswani:17} instead of LSTMs to encode inputs. Second, its training objective is masked language modelling and hence instead of predicting the next word, BERT predicts a randomly masked word from a given sentence. Third, BERT uses subword tokens instead of word tokens, thus  some infrequent words get eliminated and their common sub-parts are utilized\footnotemark\footnotetext{Word sub-parts were shown to reduce data sparsity problem in morphologically rich languages such as German~\citep{Kudo:18a}, and Turkish \citep{Akkaya:20}}. Finally, pre-trained language model can be fine-tuned for a specific language task at hand by adding one last layer on top of the utilized neural architecture. Thereafter, multilingual BERT (mBERT) was released to support many languages in a single model. However, some research studies demonstrated that BERT trained for a single language outperforms mBERT in several tasks such as dependency parsing and natural language inference~\citep{Martin:19}.  Robustly optimized BERT pretraining~\citep{Liu:19} demonstrated that longer training with careful hyperparameter selection could achieve better results as compared to earlier studies. Another transformer XLM-RoBERTa (XLMR) combined robustly optimized BERT pretraining approach with cross lingual language pretraining~\citep{Lample:19}, while using a larger dataset, and outperformed mBERT in most tasks. In other studies, transformer based architectures were both explored with and without the addition of a CRF layer. For instance, named entity recognition in Slavic languages~\citep{Arkhipov:19} and in Portuguese were confirmed to be improved once a trained BERT model is accompanied with a CRF layer \citep{Souza:19}.

\begin{table}
\caption{Turkish NER Studies}
\label{tabl:tr_lit_rev}
\begin{tabular}{lll}
\toprule Study &  Approach & F1 Score\\ 
\midrule
\citet{Kuru:16}   &  LSTM            & 91.30 \\
\citet{Demir:14} & Reg. Avg. Percp.   & 91.85 \\
\citet{Gungor:19}  & BiLSTM-CRF                           & 92.93 \\ 
\citet{Gunes:18}     &  Deep-BiLSTM       & 93.69 \\
\citet{Stefan:20}     &  BERT      & 95.40 \\
\bottomrule
\end{tabular}
\end{table}

\subsection{Turkish Named Entity Recognition}\label{subsec:TurNER}

Turkish is an agglutinative language with rather complex morphotactics where a lot of information is encoded (such as syntactic roles and relations of words) in morphology. Several Turkish words can be derived by appending multiple suffixes (i.e., inflectional and derivational) to a nominal or verbal root, as often seen in other morphologically rich languages such as Finnish, Hungarian, and Czech. The morphological structure of a Turkish word can be represented as a sequence of inflectional morphemes (IG) separated by derivation boundaries(ˆDB). Each IG sequence has its own part of speech (POS) and inflectional features, and the beginning of a new sequence is marked by a derivation boundary where a change in part of speech occurs. A word might have multiple such representations due to morphological ambiguity. For instance, the following is one possible representation of the word ``haberle\c{s}meliyiz'' (\textit{we should communicate}) where the first IG represents that the root is a verb and it is transformed into a noun with the addition of the 2$^{nd}$ infinitive suffix ``-me":

haberleş+Verb+Pos \

\hspace{0.4in}  \textasciicircum DB+Noun+Inf2+A3sg+Pnon+Nom\

\hspace{0.4in} \textasciicircum DB+Adj+With\

\hspace{0.4in} \textasciicircum DB+Noun+Zero+A3sg+Pnon+Nom\

\hspace{0.4in} \textasciicircum DB+Verb+Zero+Pres+A1pl \newline 

Although surface forms are constrained by morphological rules (e.g., vowel harmony and vowel drops) \citep{Oflazer:94}, the number of derived words from a single root is still very large to be handled easily and lexical sparsity is often experienced in learning-based NLP applications. For instance, in a Turkish dataset of 10 million words, the vocabulary size is measured as 474,957 whereas that number is lowered to 97,734 unique words in an English dataset of the same size \citep{Hakkani:00}. However, the vocabulary size is observed to degrade to 94,235 unique words once the root forms of Turkish words are considered over the same dataset. As a common practice to handle data sparsity, Turkish NLP studies often utilize disambiguated morphological representations of words rather than their surface forms.

Named Entity Recognition in Turkish has been studied for many years \citep{Kucuk:17}. The first statistical Turkish NER study \citep{Tur:03} trained an HMM model to tag person, location, and organization names that appear in well-written texts by leveraging morphological, lexical and contextual information of words. In another study \citep{Kucuk:10}, a rule-based approach was explored where knowledge resources such as dictionaries of person and location names, and pattern extraction rules for temporal and numeric expressions are heavily utilized. The system was then enriched with an ability to learn knowledge resources from annotated data. A CRF based NER system \citep{Yeniterzi:11} highlighted the impact of morphology on tagging process and benefited from roots and morphological features of words as separate tokens instead of words. An automated rule learning system \citep{Tatar:11}, a CRF based system relying on the use of gazeteer and hand crafted morphology dependent features \citep{Seker:12}, and a classification system where six different models are trained with both discrete and continuous features of words \citep{Ertopcu:17} are among recent Turkish NER studies. Although we use the same dataset for training and testing purposes \citep{Tur:03}, our work utilizes a neural network based solution and hence significantly differs from these earlier rule-based or statistical approaches. 

The first neural network based study \citep{Demir:14} used a regularized average perceptron algorithm and combined continuous vector representations of words and some language independent features (such as  context, previous tags, and case features) in a semi-supervised fashion. The use of character embeddings rather than word embeddings was later explored in a stacked bidirectional LSTM network \citep{Kuru:16}. For each input character, the system outputs a tag probability and a Viterbi decoder converts character-level probabilities to word-level tag probabilities. The results demonstrated that a good tagging performance could be achieved without benefiting from an extensive list of word features and language dependent knowledge resources. The current state-of-the-art systems utilize bidirectional LSTM networks and experiment with different word representations. The first BiLSTM study \citep{Gungor:19} concatenated word, character, and morphological embeddings as encoder inputs and used a CRF layer on top of the decoder. The tagging model was tested on four other morphologically rich languages (i.e., Czech, Hungarian, Spannish, and Finnish) and the results demonstrated that word representations once augmented with morphological and character embeddings achieve the highest performance. On the other hand, the second BiLSTM study \citep{Gunes:18} combined word embeddings, writing style embeddings (e.g., all in uppercase letters or in sentence case letters) as input representations, and experimented with stacked layers of varying depth. There is only one work where deep bidirectional transformers were utilized \citep{Stefan:20}, and in that study both cased and uncased BERT models were evaluated on Turkish NER task. The performances of these systems are listed in Table~\ref{tabl:tr_lit_rev}. Our work lies on the path opened by these BiLSTM studies where different embeddings are learned and sequentially encoded by LSTMs. However, to the best of our knowledge, this work is the first Turkish NER study that compares language models learned by transformers with BiLSTM models in the same experimental setup. Moreover, our work explores the impact of context on task performance by exploring both context sensitive and insensitive word embeddings. 

In recent years, another branch of Turkish NER studies has focused on noisy data specifically from social media. Although a limited number of approaches \citep{Celikkaya:13, Eken:15, Okur:16, Akkaya:20} have provided different solutions to the task, they all continuously  improved on the state of the art. The current state of the art with an f-score of 67.39\% is still behind the observed performances on clean data. 

\section{System Architecture} \label{sec:Architecture}

Named entity recognition is a labeling task over a text that consists of a sequence of words, and hence any approach that tags every single word with a label from a predetermined set would be a reasonable solution. In this work, we address the task as a sequence to sequence (Seq2Seq) learning problem and build two different architectures for tagging. The first architecture utilizes a Bidirectional Long Short Term Memory (BiLSTM) network whereas the second architecture uses a Transformer-based neural network. A CRF layer is employed on top of these architectures as an optimization layer for predicting the best label sequence. Our study has similarities with some earlier works~\citep{Gunes:18, Gungor:19, Akkaya:20}, but the main difference comes from the utilization of a context-sensitive language model and its performance comparisons with well-studied LSTM based language models.

\subsection{BiLSTM Network} \label{subsec:LSTM}

BiLSTM architectures utilize two separate LSTM networks \citep{Hochreiter:97}, a specialized form of recurrent neural networks that can cope with vanishing gradient problem. The first LSTM network processes input in the forward direction to keep a history from the beginning of the sequence whereas the second LSTM network processes all words in the sequence starting from the end of the input. 

  \begin{figure}[align, pos=t]
	\centering
	\includegraphics[width=0.9\linewidth]{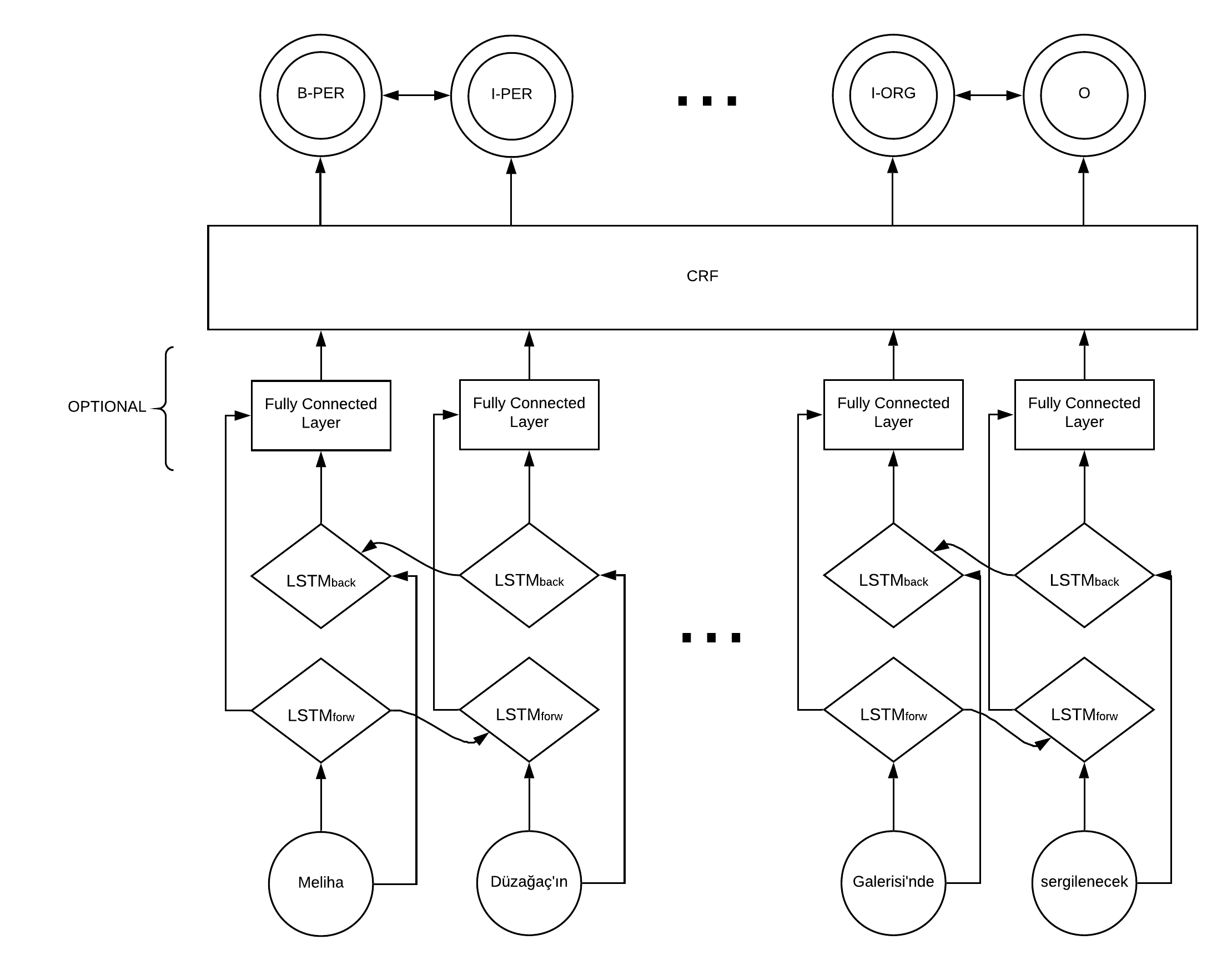}
	\caption{BiLSTM-CRF Architecture.}
	\label{fig:crf}
  \end{figure}

Our problem is formulated as given an input sentence S=\{s$_1$,s$_2$,...,s$_n$\} consisting of n words, obtain a sequence of labels L=\{l$_1$,l$_2$,...,l$_n$\} such that each l$_m$ is from a set of NER tags. As shown in Figure~\ref{fig:crf}, a word that appears in the sequence is encoded into an embedding (x$_e$) and then fed to the network. In the current implementation, each word is encoded by a combination of four different embeddings: 

\begin{itemize}
 \setlength{\parskip}{0in}
 \setlength{\itemsep}{0in}
\item Word embedding: Vector representation of the word as a token ($w_{e}$)
\item Subword embedding: Vector representation of the word chunk as a token ($sw_{e}$)
\item Character embedding: Vector representation of the word at character-level ($c_{e}$)
\item Morphological embedding: Vector representation of the word at morphological-level ($m_{e}$)
\end{itemize}

\begin{figure*}[align, pos=h] 
	\centering
    \includegraphics[width=0.38\textwidth]{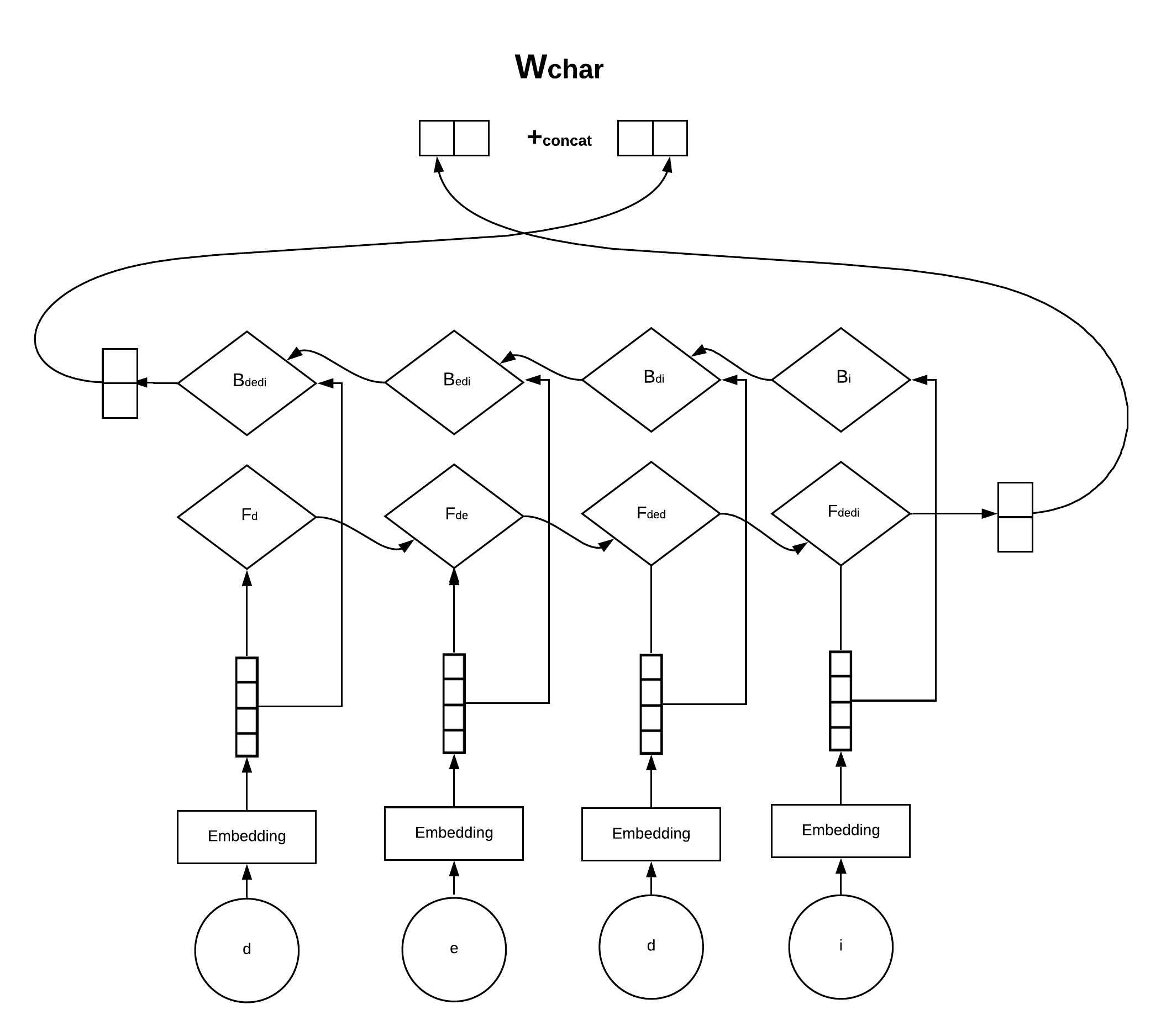} 
	\includegraphics[width=0.45\textwidth]{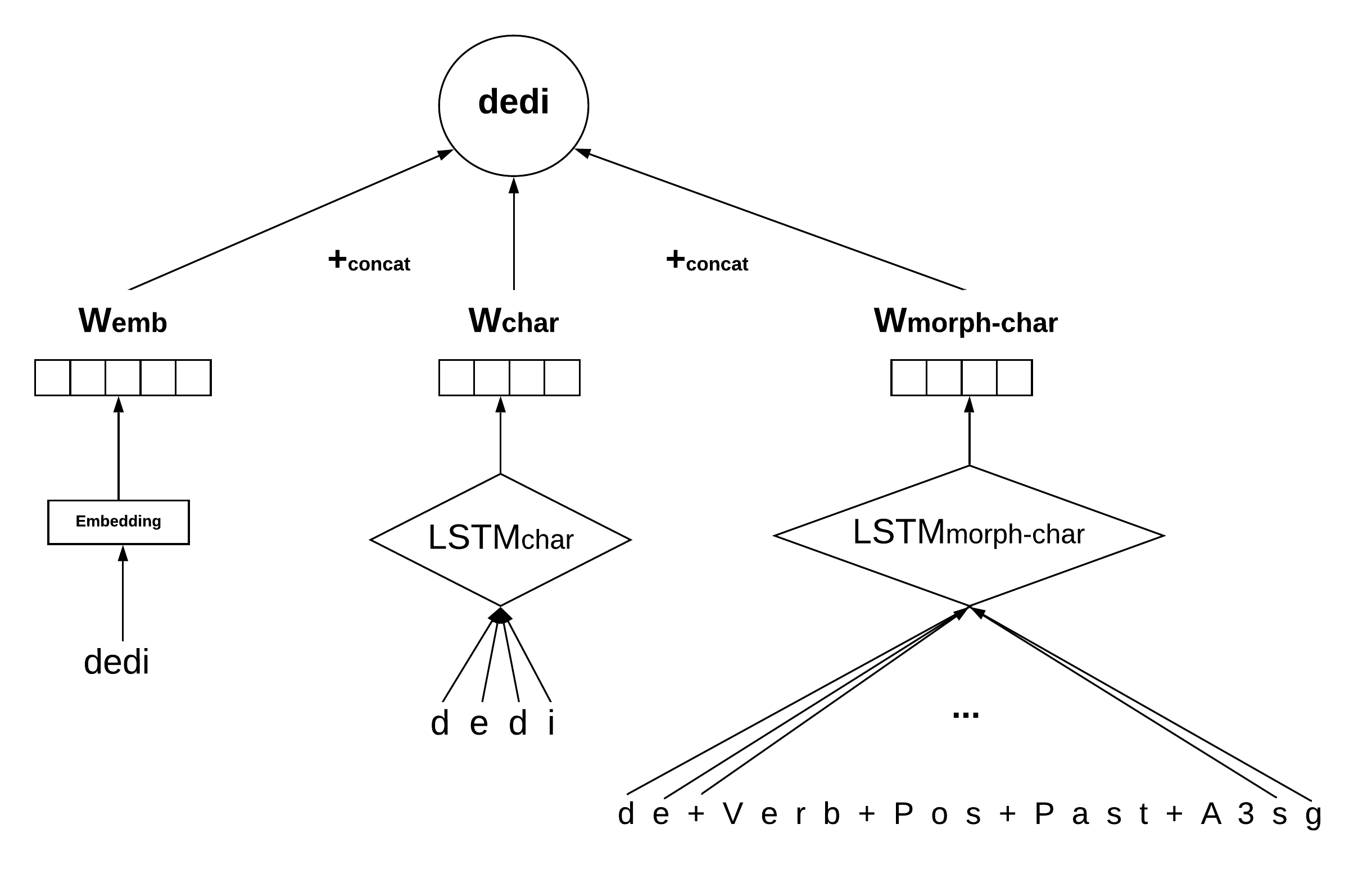}
	\caption{ a) Character Embedding  and b) Input Embedding of the Word ``dedi"(\textit{he/she said}).}
	\label{fig:word_embed}
\end{figure*}

We use a context-insensitive language model to obtain the word embedding ($w_{e}$) of each token, where every word in the sequence is taken as a single token. This embedding neither captures the location of the word in the sequence nor the contextual content of the input. We obtain subword, character, and morphological embeddings of words using distinct BiLSTM networks. For instance, the network that produces character embeddings processes every character in a word as a separate token, as shown in Figure~\ref{fig:word_embed}-a. On the other hand, morphological BiLSTM network with a similar architecture produces embeddings to reflect morphological subunit information of each word in the sequence. Subword embeddings exploit the highest possible similarity between different words. These four kinds of embeddings are utilized in order to capture the morphologically-rich nature of Turkish and information encoded in terms of characters, morphemes, and word chunks. Separate embeddings allow us to explore different ways of concatenating them to obtain the final input embedding used by our architecture. For instance, model shown in Figure~\ref{fig:word_embed}, concatenates word, character, and morphological embeddings in order to obtain the final input word embedding (i.e.,  $x_{e} = w_{e} \oplus c_{e} \oplus m_{e}$).  

The computations performed in our architecture with LSTM cells are as follows:

\begin{align}
\begin{split}
\mathbf{i}_{t} = \sigma(\mathbf{W}_{ii}\mathbf{x}_{t} + \mathbf{b}_{ii} + \mathbf{W}_{hi}\mathbf{h}_{t-1}  + \mathbf{b}_{hi}) \\
\mathbf{f}_{t} = \sigma(\mathbf{W}_{if}\mathbf{x}_{t} + \mathbf{W}_{hf}\mathbf{h}_{t-1}+ \mathbf{b}_{hf})\\
\mathbf{g}_{t} = \tanh(\mathbf{W}_{ig}\mathbf{x}_{t} + \mathbf{b}_{ig} + \mathbf{W}_{hg}\mathbf{h}_{t-1} + \mathbf{b}_{hg})\\
\mathbf{o}_{t} = \sigma(\mathbf{W}_{io}\mathbf{x}_{t} + + \mathbf{b}_{io} + \mathbf{W}_{ho}\mathbf{h}_{t-1} +  \mathbf{b}_{ho}) \\
\mathbf{c}_{t} = \mathbf{f}_{t} * \mathbf{c}_{t-1} + \mathbf{i}_{t} * \mathbf{g}_{t} \\
\mathbf{h}_{t} = \mathbf{o}_{t} * \tanh(\mathbf{c}_{t}) 
\end{split}
\end{align}

where $\mathbf{h}_{t}$ is the hidden state, $\mathbf{c}_{t}$ is the cell state, and $\mathbf{x}_{t}$ is the input at time $t$, and $\mathbf{i}_{t}$, $\mathbf{f}_{t}$, $\mathbf{g}_{t}$, and $\mathbf{o}_{t}$ are the input, forget, cell, and output gates, respectively.

\subsection{Transformer-Based Network} \label{subsec:Transform}

\begin{figure}[align, pos=h] 
	\centering
	\includegraphics[width=0.8\linewidth]{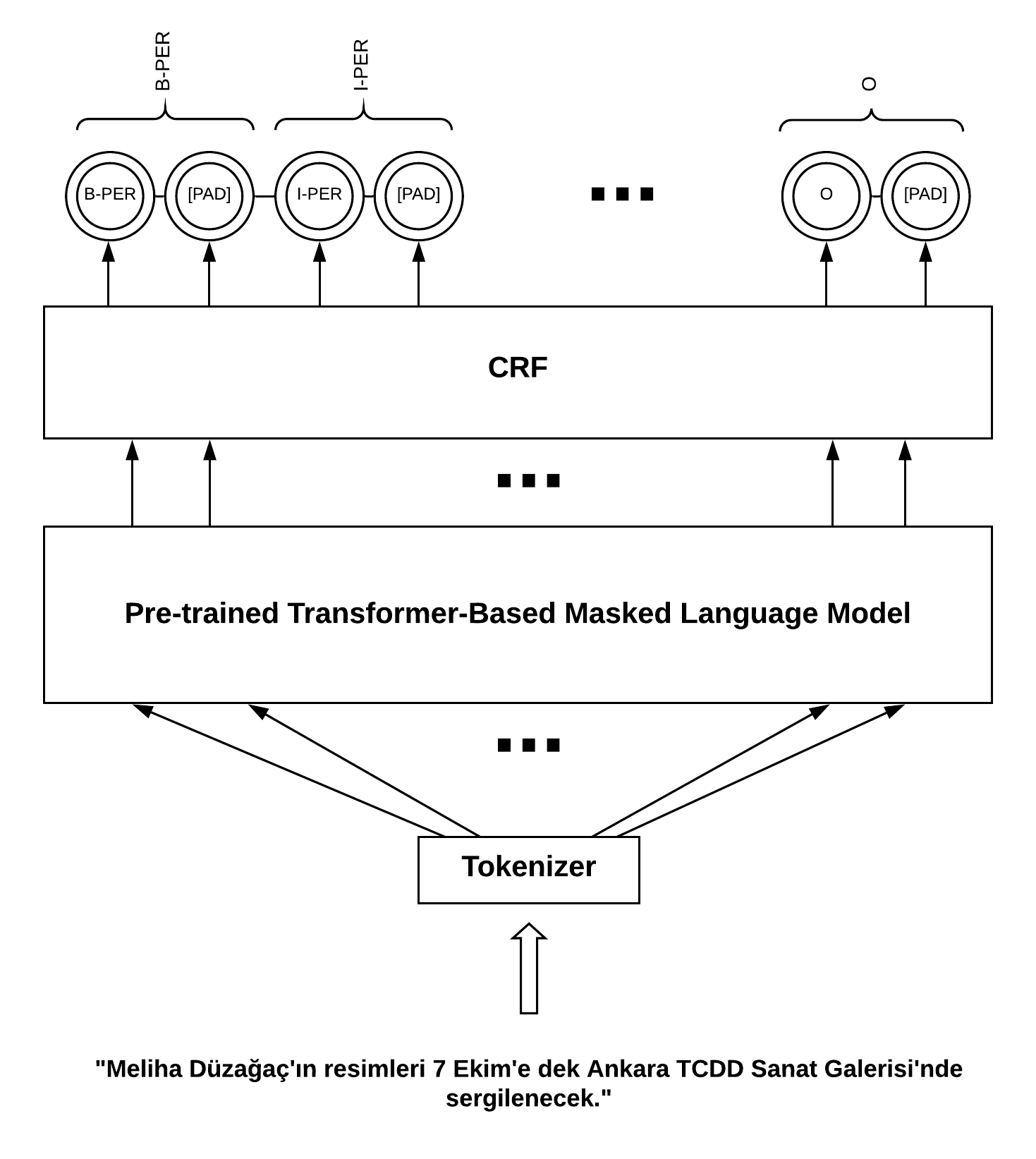}
	\caption{Transformer-Based Network.}
	\label{fig:bert-crf}    
\end{figure}

Transformer-based language models replace recurrent neural network cells with self attention and fully connected layers. As a result, the content of a whole sentence and the location of each word in the sentence are effectively captured to encode contextual information and long-range dependencies. Conditioning on both the left and right contexts of a word results in dissimilar encodings for the same word in different sentences. Moreover, these models enable us to benefit from shared embeddings between multiple natural languages and subword units in monolingual settings. In this architecture, we use pretrained masked language models and fine tune them on the NER task. As show in Figure \ref{fig:bert-crf}, the input sequence is first tokenized into subword units and then fed to the network.

\subsection{CRF Layer} \label{subsec:CRF}

The CRF layer is utilized as the top hidden layer in both architectures. This layer takes the concatenation of last hidden states from the underlying network. Its role is modeling the joint probability of the entire label sequence, in order to impose constraints over neighbour tokens \citep{Lafferty:01}. A standard implementation is carried out \citep{Zhang:19} and the probability of a label sequence $L=l_1,l_2,\ldots,l_n$ is calculated as follows:

\begin{equation}\label{eq:crf}
{
P(L|S)=\dfrac{exp(\sum_{i}(\textbf{W}_\textsc{Crf}^{l_i}\textbf{h}_i + b_\textsc{Crf}^{(l_{i-1},l_i)}))}{\sum_{L^\prime}exp(\sum_{i}(\textbf{W}_\textsc{Crf}^{l_i^\prime}\textbf{h}_i + b_\textsc{Crf}^{(l_{i-1}^\prime, l_i^\prime)}))}
}
\end{equation}

where $L^\prime$ represents an arbitrary label sequence, and $\textbf{W}_\textsc{Crf}^{l_i}$ is  a model parameter specific to $l_i$, and $b_\textsc{Crf}^{(l_{i-1},l_i)}$ is a bias specific to $l_{i-1}$ and $l_i$. 

For decoding, a first-order Viterbi algorithm is used to find the most probable  label sequence over the input sequence, and sentence-level log-likelihood loss with $L_2$ regularization is used to train the model:
\begin{align}
\begin{split}
L = \sum_{i=1}^{N}log(P(y_i|s_i)) + \frac{\lambda}{2} ||\Theta||^2 \\ 
\end{split}
\end{align}

where  $\{(s_i, l_i)\}|^N_{i=1}$ is a set of manually labeled data, $\lambda$ is the $L_2$ regularization parameter, and $\Theta$ is the parameter set.

\section{Experimental Setup} \label{sec:Experiments}
 
\subsection{Dataset}\label{subsec:preprocess}

The dataset used in this study \citep{Tur:03} is a collection of articles from the national newspaper Milliyet, covering a period between 1 January 1997 and 12 September 1998. The dataset contains Turkish sentences tagged with BIO2 scheme in CoNNL format and morphological analyses of all sentence tokens. For instance, the sentence from the dataset  ``Meliha Düzağaç'ın resimleri 7 Ekim'e dek Ankara TCDD Sanat Galerisi'nde sergilenecek." (\textit{Meliha Düzağaç's paintings will be exhibited at Ankara TCDD Arts Gallery until 7th of October.}) is tagged as follows:\newline

\texttt{\hspace{0.15in}  Meliha B-PERSON} \

\texttt{\hspace{0.2in}Düzağaç'ın I-PERSON} \

\texttt{\hspace{0.2in}resimleri O \ }

\texttt{\hspace{0.2in}7 O} \

\texttt{\hspace{0.2in}Ekim'e O} \

\texttt{\hspace{0.2in}dek O} \

\texttt{\hspace{0.2in}Ankara B-ORGANIZATION} \

\texttt{\hspace{0.2in}TCDD I-ORGANIZATION} \

\texttt{\hspace{0.2in}Sanat I-ORGANIZATION} \

\texttt{\hspace{0.2in}Galerisi'nde I-ORGANIZATION} \

\texttt{\hspace{0.2in}sergilenecek O} \

\texttt{\hspace{0.2in} . O\ }\newline

In BIO2 scheme tagging, the first token of a named entity of a particular type (type$_x$) is labeled with beginning type tag (B-type$_x$), and the remaining tokens of the same named entity are labeled with the inside type tag (I-type$_x$). All other tokens that do not belong to a named entity are labeled with outside type tag (O). In this work, we split the dataset into a training set of 32,171 sentences, 20\% of which is reserved as validation set, and a test set of 3328 sentences.

\subsection{Building BiLSTM Networks} \label{subsec:BuildBiLSTM}

\begin{table*}[align]
\caption{Sets of Word Embeddings}
\begin{tabular}{lllllll} 
\toprule
Name   & Training Method & Dataset Size  & Vocabulary Size  &  Dimension & Window Size & Negative Sampling\\ \midrule
Hur  & Skip-gram & $\sim$170M & 500 K & 128 & 1 & 2\\
Huaw  & Skip-gram  &  941 M & 1.2 M & 300 & 5 & 10        \\
FastText  & CBOW &- & 2M & 300 & 5 & 10   \\
Random  & Randomly initialized  & - &  1.2 M & 300 & - & -         \\ \bottomrule
\end{tabular}
\label{tabl:emb_settings}
\end{table*}

To generate input embeddings of the encoder, we first obtain vector representations of each token in our dataset. For character, morphological, and subword embeddings, vectors are randomly initialized whereas four different initializations are experimented in word embeddings (Table ~\ref{tabl:emb_settings}):

\begin{itemize}
 \setlength{\parskip}{0in}
 \setlength{\itemsep}{0in}
\item  \textit{Hur:} Embeddings obtained by applying word2vec skip-gram model~\citep{Mikolov:13} to articles published in the national newspaper Hurriyet from 1997 to 2019. 
\item \textit{Huaw:} Skip-gram embeddings used in a previous research study~\citep{Gungor:19} where embeddings are trained with a larger dataset and window size\footnotemark\footnotetext{Retrieved from \url{https://github.com/onurgu/linguistic-features-in-turkish-word-representations/releases}}.
\item \textit{FastText:} Embeddings trained with continuous bag of words model with position weights~\citep{Grave:18} using Common Crawl and Wikipedia dumps\footnotemark\footnotetext{Retrieved from \url{https://fasttext.cc/docs/en/crawl-vectors.html\#models}}.
\item \textit{Random:} Randomly initialized embeddings.
\end{itemize}

To form character embeddings, a random embedding is initialized for each character. Then these embeddings are fed into character-LSTM, that encodes sequences of characters. The embedding for a word derived from its characters is the concatenation of its forward and backward representations from the bidirectional character-LSTM (Figure \ref{fig:word_embed}-a).
Since our dataset also contains morphological analysis of each word, morphological embeddings are also utilized. Character-level morphological embeddings are used, as this representation was shown to work best in previous work \citep{Gungor:19}. Morphological embeddings have the same architecture as character embeddings, except they encode the full morphological analysis of the given word instead of the word itself (Figure \ref{fig:word_embed}-b). Word chunks used in subword embeddings are obtained via a unigram SentencePiece \citep{Kudo:18} tokenizer\footnote{Using \url{https://github.com/google/sentencepiece}}. The SentencePiece tokenizer is trained using a news archive of 14,995,202 tokens which consists of articles published in Hürriyet newspaper from 22 November 2018 to 22 November 2019. The unigram tokenizer, that we refer to as Turkish SentencePiece (TR SentPiece) tokenizer, has a vocabulary of size 50,000 tokens. In our experiments, we use an embedding size of 300 for words and subwords whereas 200 for characters and morphological units. Using different combinations of these embeddings, we train several BiLSTM networks using stochastic gradient descent optimizer with an initial learning rate of 0.05 (some of which are listed in Table~\ref{table:lstmscores}). In these trainings, gradient clipping of 0.5 is used and dropout is applied to concatenated embeddings with the probability of 0.5. Each models is trained for 50 epochs and 0.9 momentum is used within the optimizer. Moreover, learning rate decay is applied at the end of every epoch using the following function: \newline\newline
\indent $ lr =  lr_{previous} * (1 / (1 + 0.05 * epoch))$  
 
\subsection{Building Transformer-Based Networks} \label{subsec:BuildTrans}

To build our transformer-based networks, we utilize pretrained language models multilingual cased BERT (mBERT), Turkish BERT (BERTurk)\footnote{\url{https://huggingface.co/dbmdz/bert-base-turkish-cased}}, and XLM-RoBERTa (XLMR). For each model, we experiment with two different settings:
\begin{itemize}
 \setlength{\parskip}{0in}
 \setlength{\itemsep}{0in}
\item The model is followed with a linear layer and cross-entropy is used as loss function
\item The model is followed with a CRF layer and negative log is used as loss function 
\end{itemize}

In both settings, finetuning is applied to language models and sentences are tokenized by default tokenizers. However, in the first setting, subwords that do not appear in the first position of words are treated as padding tokens in loss calculations of training. In the evaluation phase, a BIO2 tag is assigned to only first subword token of a word and the remaining subwords of a word (treated as padding in training) are labeled with the same tag. It is worth mentioning that default tokenizers provided with language models produce different subword tokens for the same sentence. For instance, outputs produced by all tokenizers used in this study for the sentence ``Meliha Düzağaç'ın resimleri 7 Ekim'e dek Ankara TCDD Sanat Galerisi'nde sergilenecek.'' are as follows: \newline

\noindent \texttt{Morphological Analysis\footnotemark\footnotetext{ + and ++ represent inflectional and derivational suffixes, respectively.}:} \newline
\noindent\texttt{["Meliha", "Düzağaç",  "\textquoteright",  "+ın",  "resim",  "+ler",  "+i",  "7",  "Ekim", "\textquoteright",  "+e",  "dek",  "Ankara",  "TCDD",  "Sanat",  "Galeri",  "+si",  "\textquoteright",  "+n" ,  "+de",  "ser",  "++gi",  "++len",  "+ecek",  "."]}\newline

\noindent \texttt{BERTurk Tokenizer:}\newline
\texttt{["Melih", "\#\#a", "Düz", "\#\#ağaç", "\textquoteright", "ın", "resimleri", "7", "Ekim", "\textquoteright", "e", "dek", "Ankara", "TCDD", "Sanat", "Galerisi", "\textquoteright", "nde", "sergilen", "\#\#ecek", "."]}\newline

\noindent \texttt{mBERT Tokenizer:}\newline
\noindent \texttt{["Mel", "\#\#ih", "\#\#a", "D", "\#\#üz", "\#\#a", "\#\#ğa", "\#\#ç", "\textquoteright", "ın", "res", "\#\#im", "\#\#leri", "7", "Ekim", "\textquoteright", "e", "dek", "Ankara", "TC", "\#\#D", "\#\#D", "Sanat", "Gale","\#\#risi", "\textquoteright", "nde", "ser", "\#\#gile", "\#\#nec", "\#\#ek", "."]} \newline \newline

\noindent \texttt{XLMR Tokenizer:}\newline
\noindent \texttt{["\underline{\hspace{0.25cm}}Meli", "ha", "\underline{\hspace{0.25cm}}Düz", "ağa", "ç", "\textquoteright", "ın", "\underline{\hspace{0.25cm}}resim", "leri", "\underline{\hspace{0.25cm}}7", "\underline{\hspace{0.25cm}}Ekim", "\textquoteright", "e", "\underline{\hspace{0.25cm}}de", "k", "\underline{\hspace{0.25cm}}Ankara", "\underline{\hspace{0.25cm}}TC", "DD", "\underline{\hspace{0.25cm}}Sanat", "\underline{\hspace{0.25cm}}Galeri", "si", "\textquoteright", "nde", "\underline{\hspace{0.25cm}}sergi", "lenecek", "."]]}\newline

\noindent \texttt{TR SentPiece Tokenizer:}\newline
\noindent \texttt{["\underline{\hspace{0.25cm}}Melih", "a", "\underline{\hspace{0.25cm}}Düz", "ağaç", "\textquoteright", "ın", "\underline{\hspace{0.25cm}}resimleri", "\underline{\hspace{0.25cm}}7", "\underline{\hspace{0.25cm}}Ekim", "\textquoteright", "e", "\underline{\hspace{0.25cm}}dek", "\underline{\hspace{0.25cm}}Ankara", "\underline{\hspace{0.25cm}}TCDD", "\underline{\hspace{0.25cm}}Sanat", "\underline{\hspace{0.25cm}}Galerisi", "\textquoteright", "nde", "\underline{\hspace{0.25cm}}sergilenecek", "."]}\newline

In our preliminary experiments, we observe that tokens produced by multilingual BERT tokenizer do not correlate well with morphological units given in the dataset. Although this is not the case for other tokenizers, BERTurk has a small vocabulary size and XMLR is not trained solely for Turkish language. Due to these reasons, we do not report any results on the use of default tokenizers in BiLSTM networks. Moreover, in our preliminary experiments, we observe around 20\%-40\% mismatches between subword tokens obtained by our trained Tr SentPiece tokenizer and vocabularies used in pretrained models. Thus, we do not report results regarding the use of Tr SentPiece tokenizer in any of our transformer-based networks. HuggingFace transformers library\footnote{\url{https://github.com/huggingface/transformers}} with PyTorch \citep{Wolf:19} is used for implementations. Networks are trained by Adam optimizer with fixed weight decay, with initial learning rate of $5\mathrm{e}{-05}$, and gradient clipping of 1. Table \ref{table:bert_settings} provides details of all language models used in transformer-based networks.

\begin{table}[pos=h]
\caption{Settings of Masked Language Models}
\begin{center}
\begin{tabularx}{\linewidth}{@{}LL X L X@{}} 
\toprule
\multirow{2}{*}{Model}   & \# Hidden  & \# Hidden  & \# Attention  & Vocabulary \\ 
  &  Layers &  Units & Heads & Size \\ \midrule
mBERT\hspace{0.2in}   & 12                      & 768         & 12                        & 119,547         \\
BERTurk & 12                      & 768         & 12                        & 32,000          \\
XLMR-b  & 12                      & 768         & 12                        & 250,002         \\
XLMR-l  & 24                      & 1024        & 16                        & 250,002         \\ \bottomrule
\end{tabularx}
\end{center}
\label{table:bert_settings}
\end{table}

\subsection{Evaluation Metrics}\label{subsec:evalmetric}

In this study, the evaluation scores are reported using standard CoNNL\footnotemark\footnotetext{The Conference on Natural Language Learning that is organized by SIGNLL (ACL's Special Interest Group on Natural Language Learning).} precision, recall, and F1 metrics. The boundaries of all entities in a test sentence are determined by grouping tokens that form a single entity (i.e., a token sequence with B- and I-tags) and scores are computed at the entity-level. The library \verb|seqeval|\footnote{\url{https://pypi.org/project/seqeval/}} is used to compute F1 scores using the formula shown below:
\begin{equation}
 F1 = \frac{2 * (precision * recall)}{(precision + recall)}
\end{equation}


\section{Results and Discussion} \label{sec:Results}

\begin{table*}[align]
\caption{Performance Scores of BiLSTM-CRF Models on Our Validation and Test Sets}
\medskip
\small
\begin{tabular}{L llllllll l}     
\toprule
Model \# & Model Description     & Embedding &Valid$\setminus$Test & F1 & Precision & Recall & Accuracy & Trn. Time \\ \midrule
\multirow{2}{*}{1}&\multirow{2}{*}{Word-Char-BiLSTM-CRF}           & \multirow{2}{*}{Random}     & Valid  &  85.75 &      84.41 &   87.15 &     97.84 & \multirow{2}{*}{11:08:25} \\
&        &     & Test  &  85.20 &      84.10 &   86.33 &     97.76&  \\\hline
\multirow{2}{*}{2}&\multirow{2}{*}{Word-Char-BiLSTM}              & \multirow{2}{*}{Huaw}    & Valid  &  86.08 &      83.59 &   88.72 &     98.21 & \multirow{2}{*}{02:07:40} \\
&        &     & Test  & 85.28 &      83.06 &   87.62 &     98.16  &  \\\hline
\multirow{2}{*}{3}&\multirow{2}{*}{Subword-Char-BiLSTM-CRF}   & \multirow{2}{*}{Random} &Valid &  86.26 &      85.29 &   87.25 &     97.09 &   \multirow{2}{*}{06:28:04} \\
&        &     & Test  &   86.37 &      84.93 &   87.85 &     97.10     &  \\\hline
\multirow{2}{*}{4}&\multirow{2}{*}{Word-Char-BiLSTM-CRF}           & \multirow{2}{*}{Hur}  &   Valid  &  87.21 &      86.14 &   88.19 &     98.04 & \multirow{2}{*}{01:59:36 }\\
&        &     & Test  &  87.92 &      87.30 &   88.56 &     98.09  &  \\\hline
\multirow{2}{*}{5}&\multirow{2}{*}{Word-BiLSTM-CRF}               & \multirow{2}{*}{Huaw}  &  Valid   &  89.10 &      89.77 &   88.44 &     98.36 & \multirow{2}{*}{01:12:20} \\
&        &     & Test  &   88.70 &      89.70 &   87.73 &     98.26 &  \\\hline
\multirow{2}{*}{6}&\multirow{2}{*}{Word-Char-BiLSTM-CRF}           & \multirow{2}{*}{FastText}  &  Valid     &  89.44 &      88.36 &   90.55 &     98.38 & \multirow{2}{*}{02:15:58}\\
&        &     & Test  &   89.99 &      89.39 &   90.60 &     98.41 & \\\hline
\multirow{2}{*}{7}&\multirow{2}{*}{Word-Char-Morph-BiLSTM-CRF} & \multirow{2}{*}{Huaw}   & Valid  &  91.52 &      90.58 &   92.48 &     98.70 & \multirow{2}{*}{05:17:29}\\
&        &     & Test  & 91.65 &      91.38 &   91.92 &     98.71  &  \\\hline
\multirow{2}{*}{8}&\multirow{2}{*}{Word-Char-BiLSTM-CRF}           & \multirow{2}{*}{Huaw}   &  Valid &  91.57 &      90.64 &   92.52 &     98.72 & \multirow{2}{*}{02:00:39} \\
&        &     & Test  &  91.84 &      91.17 &   92.52 &     98.78  &  \\
\bottomrule
\end{tabular}
\label{table:lstmscores}
\end{table*}

The literature on Turkish NER studies has benefited from BiLSTM neural networks and transformer-based networks on different settings. Although a dataset is common to all these studies, they have various similar and dissimilar design considerations, parameter settings, and initializations in their architectures. In this work, we not only provide the most comprehensive performance evaluations that compare two different architectures on the same experimental setup but also report the impact of some design choices that have not been explored before in these architectures. Following an ablation study, we present our findings and quantify the strength of effect of each design consideration in focus on different architectures. Finally, we contribute to the literature by introducing a transformed-based model with a CRF layer at the top and demonstrate that this model outperforms the current state-of-art Turkish NER studies.

\subsection{Experiments on BiLSTM CRF Networks} \label{subsec:bilstm-evals}

We built several BiLSTM models using different configurations and conducted experiments to assess the impact of a single design parameter at each turn. Table~\ref{table:lstmscores} presents the performance scores of some models (in increasing order of F1 scores) on validation and test sets, respectively. We choose these models since they reflect the general tendency of varying parameters between models. 

It is our observation that previous Turkish research has spent tremendous effort to find the best way of forming input word embeddings and explored different combinations of vectors that represent word tokens from different perspectives. Character and morphological embeddings were shown to have a positive effect on the performance of BiLSTM networks \citep{Gungor:19, Akkaya:20}. However, to the best of our knowledge, no previous research has measured the contribution of subword information in encoding. A character sequence of a word is often longer than its subword sequence and longer sequences present significant modeling challenges for Seq2Seq models. Moreover, subword representations result in modest vocabulary size and have the potential to form basis for robust feature representations once accompanied with character-based representations. Thus, we argue that unexplored effect of subword representations on BiLSTM performance is worth studying. 

\begin{figure}[pos=t]
	\centering
	\includegraphics[width=0.9\linewidth]{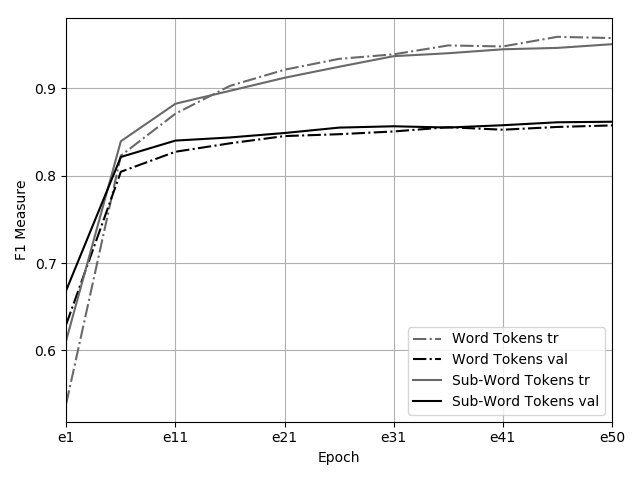}
	\caption{Performance Comparisons of Word and Subword Embeddings During Training and Validation.}
	\label{fig:f1_subw}
\end{figure}

The comparison between Model 1 and Model 3 shows a slight performance increase of 0.51\% on the validation set and 1.17\% on the test set when subword embeddings are used instead of word embeddings. The observed increase might stem from a shorter vocabulary size (reduced data sparsity) which circumvents the problem of out-of-vocabulary words up to a level. However, the rise is not that significant as we expected. This might be due to the presence of character embeddings which might efficiently encode information carried in suffixes and thus surpass advantages of subword units. Although average scores over 5 different runs are reported, one particular reason for performance differences might be the fact that randomly initialized word or subword embeddings converge differently at each run. Additionally, performance differences are observed on training and validation sets over different epochs as shown in Figure \ref{fig:f1_subw}.  

Our second set of experiments are designed to assess the contribution of word embeddings, in particular initializations of word embeddings, to tagging performance. Model 1 that uses randomly initialized word vectors achieves an f1 measure of 85.75\% on our validation set. Although, Hur and FastText pretrained embeddings both contribute to that performance with 1.46\% and 3.69\% increases respectively, the highest addition of 5.82\% comes from Huaw embeddings. We also observe similar performance improvements on the test set as shown in Table~\ref{table:lstmscores}. The results that we obtain from Models 4, 6, and 8 as compared to Model 1 motivate the need for pretrained embeddings as a good starting point. Moreover, a bigger dataset and larger word embeddings result in substantial improvements on measured performance. Moving from Hur embeddings (Model 4) to Huaw word embeddings (Model 8) provide an increase of 4.36\% on the validation set and 3.92\% on the test set. An increase of 2.23\%  on validation and 2,07\% on test sets are observed when we shift from Hur embeddings (Model 4) to FastText embeddings (Model 6) and we relate this change to dimensional differences between these embeddings. Huaw embeddings (Model 8) improve f1 scores by 2.13\% on validation and 1.85\% on test sets as compared to FastText embeddings (Model 6). This is possibly due to different methods utilized in learning representations since FastText treats each word as a composition of character ngrams, whereas Huaw embeddings are obtained by treating each word as a single token. 

\begin{figure}[pos=t]
	\centering
	\includegraphics[width=0.9\linewidth]{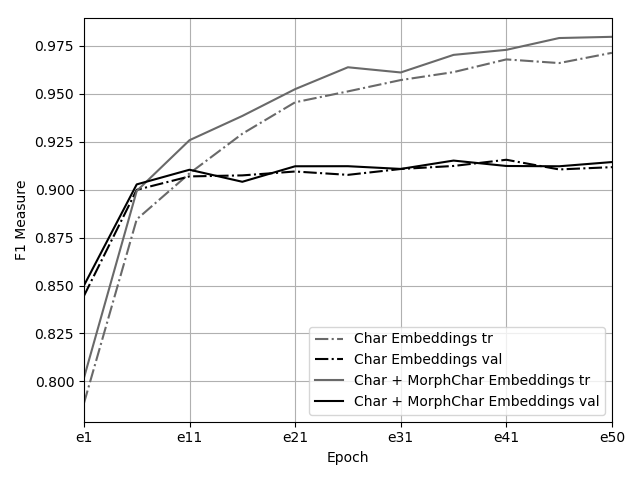}
	\caption{Performance Comparisons of Character and Character-Morphological Embeddings During Training and Validation.} 
	\label{fig:f1_char_morph_char}
\end{figure}

Our final set of experiments, in line with previous research, also confirms that the use of a CRF layer on top of the underlying architecture significantly improves f1 measures both on validation and test sets (Models 2 and 8). In a sequence labeling task, it is not surprising to see a positive effect of modeling dependencies between consecutive input tokens. In addition, f1 score obtained from Model 5 by utilizing a CRF layer is higher than that obtained from Model 2 where character embeddings are used. However, in our experiments we do not measure any notable improvements once morphological embeddings are incorporated (Models 7 and 8) and this does not support the findings of \cite{Gungor:19} where a higher improvement is obtained with the addition of morphological information. As shown in Figure \ref{fig:f1_char_morph_char} performances of these two models are very similar during training and validation. One particular reason for this divergence might be the fact that in that previous study, a character-only model was not used with a larger dimension as we did in our work. 

This experimental study reveals that the learning method used to obtain word embeddings matters, so do their dimensions; which is supported by the work of \cite{Melamud:16}. The importance of embeddings is also mentioned in the work of \cite{Ma:16} where GloVe embeddings lead to the highest performance on English. Finally, our experiments strengthen the effectiveness of utilizing character embeddings as demonstrated in the work of  \cite{Kuru:16}. This finding might be attributed to morphological information that may possibly be carried by characters.

\subsection{Experiments on Transformer-Based Networks}
\begin{table*}[align]
\caption{Performance Scores of Transformer-Based Models on Our Validation and Test Sets}
\footnotesize
*For XLMR-large, reported training time is with a larger instance type, C5.9xlarge. All other models are trained with C5.4xlarge instance type on AWS Elastic Compute Cloud service \footnote{All training was done using machines from Amazon Web Services with instance type C5.4xlarge (\url{https://aws.amazon.com/ec2/instance-types/c5/}). These machines have Intel Xeon Scalable Processors (Cascade Lake) with a sustained all-core Turbo CPU frequency of 3.6GHz, 16 vCPUs and 32 GiB of memory.}. Results are averaged over 5 random initializations.
\medskip
\small
\label{table:transformerscores}
\begin{tabularx}{\textwidth}{@{}L lllllll@{}} 
\toprule
Model \#& Model Description &   Valid.$\setminus$Test &   F1 &  Precision &    Recall &  Accuracy &   Trn. Time \\
\midrule
\multirow{2}{*}{1}& \multirow{2}{*}{mBERT-CRF}   &Valid &  92.65 &      91.90 &   93.42 &     98.92 & 03:47:00 \\
 & &Test &  92.35 &      91.54 &   93.17 &     98.90 \\\hline
\multirow{2}{*}{2}&\multirow{2}{*}{mBERT}      &Valid &  92.73 &      92.07 &   93.40 &     98.96 & 01:48:25 \\
 &    &Test &  92.59 &      91.74 &   93.45 &     98.94  \\\hline
\multirow{2}{*}{3}&     \multirow{2}{*}{XLMR-b-CRF} &Valid &  93.29 &      92.65 &   93.95 &     99.02 & 03:52:08 \\
&&Test &  93.89 &      93.10 &   94.69 &     99.11  \\\hline
  \multirow{2}{*}{4}& \multirow{2}{*}{XLMR-b}   &Valid   &  93.29 &      92.61 &   93.99 &     99.05 & 01:55:42 \\
 &  &Test    &  94.01 &      93.10 &   94.93 &     99.15   \\\hline
 \multirow{2}{*}{5}&   \multirow{2}{*}{XLMR-l*}   &Valid    &  94.56 &      93.90 &   95.24 &     99.21 &        03:21:20 \\
&   &Test  &  94.82 &      93.99 &   95.66 &     99.28         \\\hline
\multirow{2}{*}{6}&\multirow{2}{*}{BERTurk}     &Valid &  94.87 &      94.37 &   95.38 &     99.28 & 01:39:42 \\
 &  &Test &   95.75 &      95.41 &   96.10 &     99.41   \\\hline
\multirow{2}{*}{7}&\multirow{2}{*}{BERTurk-CRF} &Valid&  94.90 &      94.48 &   95.33 &     99.28 & 03:44:16 \\
 &&Test &  95.95 &      95.60 &   96.31 &     99.42 \\

\bottomrule
\end{tabularx}
\end{table*}

In literature, there exists only one work where a transformer-based language model, in particular BERT, was applied to NER task and that work was shown to outperform the state-of-the-art results obtained by BiLSTM networks (as shown in Table~\ref{tabl:tr_lit_rev}). However, there are a few other transformer-based large language models whose performances have not been reported for Turkish tagging task. Additionally, to our best knowledge, the impact of CRF on such models has not been evaluated before. Thus, our experiments on transformer-based networks are oriented around these research questions.

For this set of experiments, we trained three different networks where a multilingual cased BERT language model was used with and without a CRF layer at the top in the first architecture. Similarly, XLM-RoBERTa (XLMR) and Turkish BERT (BERTurk) language models were utilized along with CRF layers in the second and third networks, respectively. The results of these experiments on validation and test sets are reported in Table~\ref{table:transformerscores}. 

Our first observation reveals that mBERT (Models 1 and 2) performs comparably poorer than other models and BERTurk (Models 6 and 7) obtains highest f1 scores on both datasets. The results are as we expected for XLMR models; a higher performance is obtained once XLMR large (Model 5) is used rather than XLMR base (Model 4) with less number of hidden units and layers. Comparing multilingual models, we find that XLMR-b performs better than mBERT (with 0.56\%  and 1.42\% increases on validation and test datasets), and XLMR-l enhances this improvement by an additional rise of 1.27\% on validation set and 0.81\% on test set. In the lirerature, XLMR was shown to improve NER benchmarks in multiple languages \citep{Conneau:19}, and our findings provide additional support by showing that Turkish language is better represented with XLMR than mBERT. Some of this difference might be attributed to better subword token production by XLMR. Moreover, XLMR tokenizer produces more similar tokens to BERTurk tokenizer and uses a larger model settings and corpus. We argue that, due to these reasons, it achieves a closer performance to BERTurk (0.93\% difference on test set) than mBERT (2.23\% difference on test set). 

It is quite surprising to measure lower performances in mBERT (Model 1) and XLMR (Model 3) models when it is accompanied with a CRF layer. However, CRF on BERTurk (Model 7) slightly improves f1 scores with an increase of 0.03\% on valid and 0.2\% on test sets (as compared to Model 6), respectively. Although none of these performance changes are significant, our results correlate with previous studies that perform well without using a CRF layer \citep{Devlin:18, Conneau:19}. 

\begin{figure}[pos=h]
	\centering
	\includegraphics[width=0.9\linewidth]{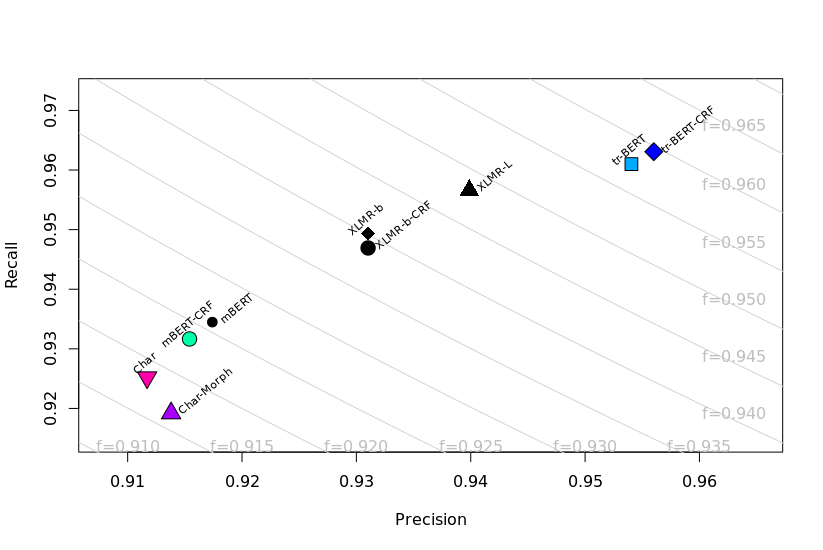}
	\caption{Performance Comparisons of BiLSTM and Transformer-Based Models on Test Set}
	\label{fig:modelcomparisons}
\end{figure}

Our final and most important finding is that all transformer-based models outperform BiLSTM models on Turkish NER task as shown in Figure~\ref{fig:modelcomparisons}. The comparison between Model 8 from Table~\ref{table:lstmscores} and Model 1 from Table~\ref{table:transformerscores} shows that at least an increase of 1.08\% on valid set and 0.54\% on test set. That improvement achieved with BERTurk-CRF (Model 7 in Table~\ref{table:transformerscores}) is at most 3.33\% on valid set and 4.11\% on test set, respectively\footnotemark\footnotetext{One particular disadvantage of transformer-based models is the observed slow inference time (between 98-211 seconds) as compared to BiLSTM models (between 8-13 seconds).}.

\section{Conclusion and Future Work} \label{sec:Conclusion}
Recent years have witnessed a surge of interests in Turkish named entity recognition. This study presents our empirical evaluations of recent neural sequence tagging models on Turkish NER task by providing a high-level comparison of different model settings and design considerations. Our results provide insights into the importance of word representations (i.e., character, morphological, subword, and word embeddings) and their initializations (i.e., random or pretrained initializations) in BiLSTM networks. Our experiments also include a comprehensive evaluation of neural architectures that utilize popular multilingual transformer-based language models on Turkish entity tagging. Their comparisons with BiLSTM models reveal their superior performance on the evaluation set and highlight the positive impact of transfer learning. In this work, we also propose a state-of-the-art transformer-based architecture with a CRF layer that achieves the highest f-measure of 94.90\% and 95.95\% on the validation and test sets, respectively. 

As our future work, we plan to aggregate character and morphological embeddings with transformer-based language models and assess their impact on the overall performance. We also intend to study other word embeddings in BiLSTM networks, especially those that were shown to be effective in other morphologically rich languages such as Flair \citep{Akbik:18}. Finally, we plan to develop new subword tokenizers such as a tokenizer that returns morphemes attached to a word as produced by a morphological analyzer. 

\bibliographystyle{cas-model2-names}

\bibliography{bib}





\end{document}